\documentclass{article}
\usepackage{spconf,amsmath,graphicx}
\usepackage{amssymb}
\usepackage{algorithm}  
\usepackage{algorithmicx}  
\usepackage{algpseudocode}  
\usepackage{amsmath}  
\usepackage{bbding}

\ninept

\title{RoIMix: Proposal-Fusion among Multiple Images \\ 
for Underwater Object Detection}
\name{Wei-Hong Lin{$^{1,2}$}, Jia-Xing Zhong{$^{1,2}$}, Shan Liu{$^{3}$}, Thomas Li{$^{4}$}, Ge Li{$^{*1,2}$}\thanks{{$^*$}Corresponding Author. ~~~This project was supported by Shenzhen Municipal Science and Technology Program (No.JCYJ20170818141146428), National Engineering Laboratory for Video Technology - Shenzhen Division, National Natural Science Foundation of China and Guangdong Province Scientific Research on Big Data (No.U1611461).}}
\address{{$^1$}School of Electronic and Computer Engineering, Peking University Shenzhen Graduate School \\
{$^2$}Peng Cheng Laboratory ~~~ {$^3$}Tencent America \\ {$^4$}Advanced Institute of Information Technology, Peking University}

\begin{document}
%
\maketitle
\begin{abstract}
Generic object detection algorithms have proven their excellent performance in recent years. However, object detection on underwater datasets is still less explored. In contrast to generic datasets, underwater images usually have color shift and low contrast; sediment would cause blurring in underwater images. In addition, underwater creatures often appear closely to each other on images due to their living habits. To address these issues, our work investigates augmentation policies to simulate overlapping, occluded and blurred objects, and we construct a model capable of achieving better generalization. We propose an augmentation method called RoIMix, which characterizes interactions among images. Proposals extracted from different images are mixed together. Previous data augmentation methods operate on a single image while we apply RoIMix to multiple images to create enhanced samples as training data. Experiments show that our proposed method improves the performance of region-based object detectors on both Pascal VOC and URPC datasets. 
\end{abstract}
\section{Introduction}
\label{sec:intro}

\begin{figure}[htb]
\begin{minipage}[b]{.48\linewidth}
  \centering
  \centerline{\includegraphics[width=4.0cm, height=1.94cm]{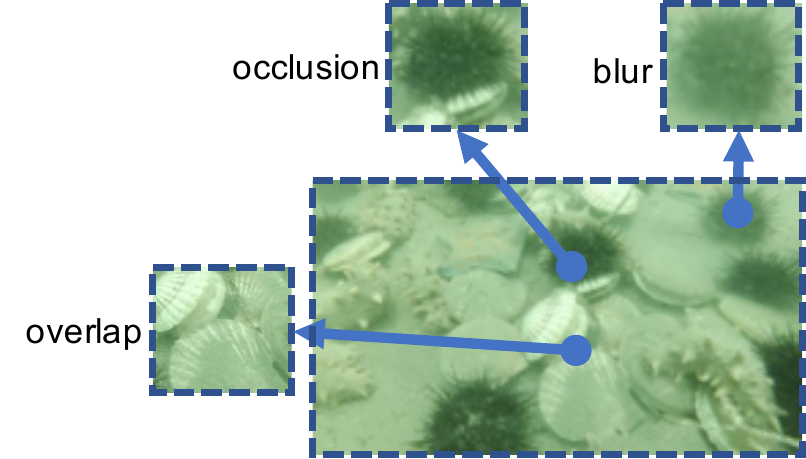}}
  \centerline{(a)}\medskip
\end{minipage}
\hfill
\begin{minipage}[b]{0.48\linewidth}
  \centering
  \centerline{\includegraphics[width=3.0cm, height=1.94cm]{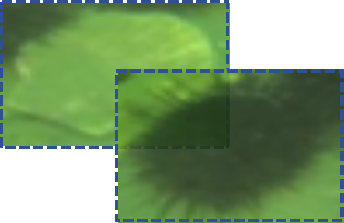}}
  \centerline{(b) }\medskip
\end{minipage}
\caption{(a) Examples of overlap, occlusion and blur. In this paper, ``overlap'' means the objects of the same class cover part of each other whereas ``occlusion'' represents the similar case for different classes. ``blur'' is caused by sediment. (b) Misaligned proposals.}
\label{fig:dataset}
\end{figure}

Many object detectors \cite{ren2015faster,cai2018cascade,pang2019libra} achieve promising performance on generic datasets such as Pascal VOC \cite{everingham2010pascal}, MSCOCO \cite{lin2014microsoft}. However, due to the complicated underwater environment and illumination conditions, underwater images often have low contrast, texture distortion and uneven illumination. Figure \ref{fig:dataset}(a) displays densely distributed creatures. They cover each other, and some of them are blurred due to sediment. These object detectors perform well on generic datasets while their capability of detecting underwater objects have been less studied. Underwater Robot Picking Contest (URPC) \footnote{http://en.cnurpc.org/} offers a challenging object detection dataset, which contains a wide range of overlapping, occluded and blurred underwater creatures.

The issue of overlapping, occluded, and blurred objects has not been well researched under the existing data augmentation methods \cite{krizhevsky2012imagenet,simonyan2014very}. If a model simply fits the existing training data, it will lack generalization ability and cannot cope with complicated underwater environment. Therefore, we directly simulate objects' overlap, occlusion and blur by mixing proposals among multiple images. 

Theoretically, following the Empirical Risk Minimization (ERM) Principle \cite{vapnik1992principles}, deep models are dedicated to minimizing their average error over the training data. However, they are usually affected by  over-fitting. Specifically, ERM guides the deep models to memorize the training data rather than generalize from them. Meanwhile, these models are easily attacked by adversarial samples \cite{zhang2017mixup}. Data augmentation is utilized to resolve over-fitting. According to the Vicinal Risk Minimization (VRM) Principle \cite{chapelle2001vicinal}, the models are optimized on samples similar to training data via augmentation strategies. In the image classification domain, translating and flipping are commonly used strategies to increase the amount of training data. Some works such as Mixup \cite{zhang2017mixup}, CutMix \cite{yun2019cutmix} are devoted to creating better training data. We investigate the effect of deploying data augmentation in training object detectors.

In this work, we aim to design a new data augmentation method for underwater object detection. Though data augmentation methods \cite{devries2017improved,summers2019improved,guo2019mixup} for image classification can bring performance gain, they are not specifically designed for underwater object detectors. We propose a data augmentation method called RoIMix to improve the capability of detecting overlapping, occluded and blurred objects. Our proposed method is designed for region-based detectors such as Faster RCNN \cite{ren2015faster} and its variants \cite{dai2016r,he2017mask,lin2017feature}. In contrast to previous data augmentation methods that operate on a single image, our proposed method pays more attention to interactions among images. Applying image-level fusion like Mixup \cite{guo2019mixup} directly in object detection would cause proposals from different images misaligned, shown in Figure \ref{fig:dataset}(b). In order to accurately simulate the situations of overlap, occlusion and blur, we perform proposal-level fusion. In this way, we achieve remarkable performance for object detection on Pascal VOC and URPC datasets.

In summary, the main contributions of this paper are as follows: (1) to the best of our knowledge, this is the first work to utilize a proposal-level mix strategy to improve the performance of object detectors, especially for overlapping, occluded and blurred underwater creatures detection; (2) unlike previous data augmentation methods that process on a single image, RoIMix focuses on interactions between images and mixes proposals among multiple images; (3) our proposed method achieves remarkable performance for object detection on both URPC and Pascal VOC. Notably, we won the first prize with our proposed method RoIMix for offline target recognition in URPC 2019.

\section{RELATED WORK}
\label{sec:related}

\subsection{Data Augmentation}
\label{ssec:subhead}

Data augmentation is a critical strategy for training deep learning models. In the image classification domain, commonly used data augmentation strategies include rotation, translation \cite{cirecsan2012multi,wan2013regularization,sato2015apac} or flip. Besides, there are some works on creating better data augmentation strategies \cite{baird1992document,simard2003best,krizhevsky2012imagenet}. Zhang et al. \cite{zhang2017mixup} proposes to mix two random training images to produce vicinal training data as a regularization approach. Regional dropout methods such as Cutout \cite{devries2017improved}, erases random regions out of the input. This helps the model attend to the most discriminative part of the objects, but it would result in the loss of information. Moreover, an advanced version CutMix \cite{yun2019cutmix} cuts and pastes the patches among the training data, which greatly improves the model robustness against input corruption. For object detection, the detector adopts multiple augmentation strategies, such as photo metric distortion \cite{liu2016ssd}, image mirror \cite{girshick2018detectron} and multi-scale training \cite{chen2019mmdetection}. Apart from this, a pretrained model based on CutMix can achieve performance gain on Pascal VOC, but it is not specifically designed for object detectors. We fully consider the characteristics of the region-based detectors and propose a new data augmentation method.

\subsection{Faster R-CNN and its variants}
\label{ssec:subhead}

Faster R-CNN \cite{ren2015faster} is a milestone in the development of the two-stage object detectors. It is composed of three modules: a Head network responsible for extracting features, such as AlexNet \cite{krizhevsky2012imagenet}, VGG \cite{simonyan2014very} and ResNet \cite{he2016deep}, RPN \cite{ren2015faster}, a fully convolutional network that generates a set of region proposals on the feature map, and a RoI classifier \cite{girshick2018detectron}, making predictions for these region proposals. However, the computation is not shared in the region classification step. Dai et al. \cite{dai2016r} proposes Region-based Fully Convolutional Networks (R-FCN), extracting spatial-aware region features. It shares the computation in the classification step via removing the fully connected layer without decline in performance. Another issue with Faster R-CNN is that it uses the output features from the last layer to make predictions, which lacks the capability of detecting small objects. Therefore, Lin et al. \cite{lin2017feature} proposes Feature Pyramid Networks (FPN), which combines hierarchical features to make better predictions. In recent years, there have been other variants of Faster R-CNN \cite{cai2018cascade,chen2019hybrid,he2019improving,Guan2018Multi,wang2019region}. Our method is potentially versatile and can be applied to two-stage detectors.

\begin{table*}[h!]
\centering
 \setlength{\tabcolsep}{1mm} {
 \begin{tabular}{c | c | c c c c c | c c c c} 
 \hline
 Method & mAP  & Single & Multiple & GT & RoI & Max & holothurian & echinus & scallop & starfish\\ 
 \hline\hline
 Baseline & 73.74 & - & - & - & - & - & 72.16 & \textbf{86.95} & 52.87 & 83.00 \\
 \textbf{Proposed (RoIMix)} & \textbf{74.92} & & \checkmark & & \checkmark & \checkmark & \textbf{73.27} & 86.80 & \textbf{55.97} & 83.65 \\
 \hline
 GTMix & 74.17 & & \checkmark & \checkmark & & \checkmark & 72.30 & 86.76 & 54.68 & 82.95\\
 Single\_GTMix & 74.23 & \checkmark & & \checkmark & & \checkmark &  71.51 & 86.66 & 54.67 & \textbf{84.09}  \\
 Single\_RoIMix & 74.51 & \checkmark & & & \checkmark & \checkmark &  73.13 & 86.59 & 54.60 & 83.71\\
 RoIMix ($\mathbf{w/o}\ \mathbf{max}$) & 72.86 & & \checkmark & & \checkmark && 71.79 & 86.11 & 50.07 & 83.46\\
 Single\_RoIMix ($\mathbf{w/o}\ \mathbf{max}$) & 73.12 & \checkmark & & &\checkmark&  & 71.22 & 86.43 & 51.26 & 83.56\\
 
 \hline
 \end{tabular}}
 \caption{Detection results on the URPC 2018. ``Single'' or ``Multiple'' means applying mixing operation on a single image or multiple images; ``GT'' or ``RoI'' represents GT-wise or RoI-wise fusion; ``Max'' means performing max function when choosing mixing ratio $\lambda'$.}
 \label{table:urpc}
\end{table*}

\section{METHODOLOGY}
\label{sec:method}

\begin{figure}[htb]

\begin{minipage}[b]{1.0\linewidth}
  \centering
  \centerline{\includegraphics[width=8.4cm, height=4cm]{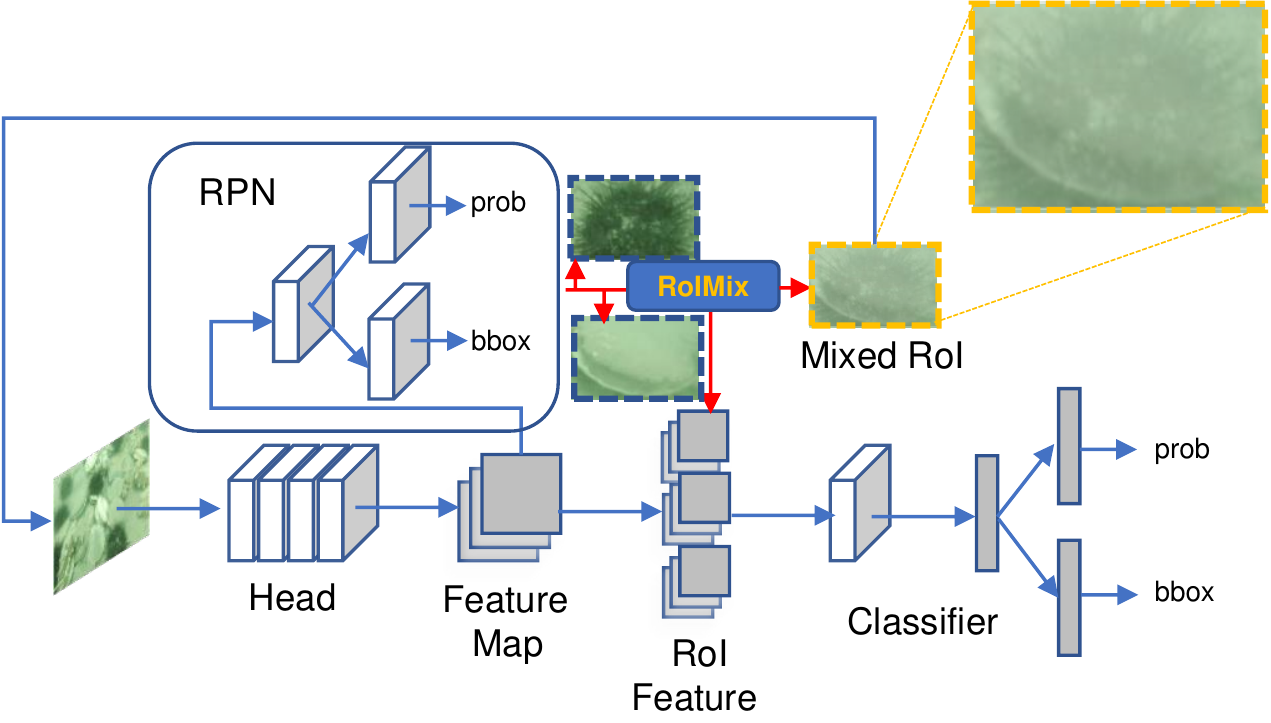}}
\end{minipage}
\caption{Overview of our approach. The architecture contains three modules: Head network, Regional Proposal Network (RPN) and Classifier. RoIMix exists between RPN and Classifier, aiming to combine random proposals generated by RPN to generate Mixed Region of Interest (Mixed RoI), and extracting the feature map of the RoIMixed Samples using for localization and classification.}
\label{fig:architecture}
\end{figure}

As shown in Figure \ref{fig:architecture}, our proposed method is applied between RPN and RoI Classifier. We take RoIs produced by RPN and mix them by a random weight ratio. The ratio is generated based on beta distribution. Then we use the mixed samples to train the model. In the following section, we will describe the algorithm of RoIMix in detail and discuss the principles behind it.

\subsection{Algorithm}
\label{ssec:subhead}

Let $x\in \mathbf{R}^{H\times W \times C}$ and $y$ denote a proposal and its label. RoIMix aims to generate virtual proposals $(\widetilde{x}, \widetilde{y})$ by combining two random RoIs $(x_i, y_i)$ and $(x_j, y_j)$ extracted from multiple images. The size of RoIs is often inconsistent, so we first resize $x_j$ to the same size as $x_i$. The generated training sample $(\widetilde{x}, \widetilde{y})$ is used to train the model. The combining operation is defined as:
\begin{equation}
    \widetilde{x}=\lambda' x_i + (1-\lambda') x_j, \widetilde{y} = y_i,
\end{equation}
where $\lambda'$ is a mixing ratio of two proposals. Instead of choosing a mixing ratio $\lambda$ directly from a Beta distribution $\mathcal{B}$ with parameter $a$ like Mixup:
\begin{equation}
    \lambda = \mathcal{B}(a, a),
\end{equation}
we pick the larger one for the first RoI $x_i$:
\begin{equation}
    \lambda' = max(\lambda, 1-\lambda),
\end{equation}
where $max$ is a function returning the larger value. The reason behind this is that we use $y_i$ as the label of mixed proposal $\widetilde{x}$. Our proposed method mixes proposals without labels, which is similar to traditional data augmentation method. It only affects training and keeps the model unchanged during evaluation.

\begin{figure}[htb]

\begin{minipage}[b]{1.0\linewidth}
  \centering
  \centerline{\includegraphics[width=8.4cm, height=2cm]{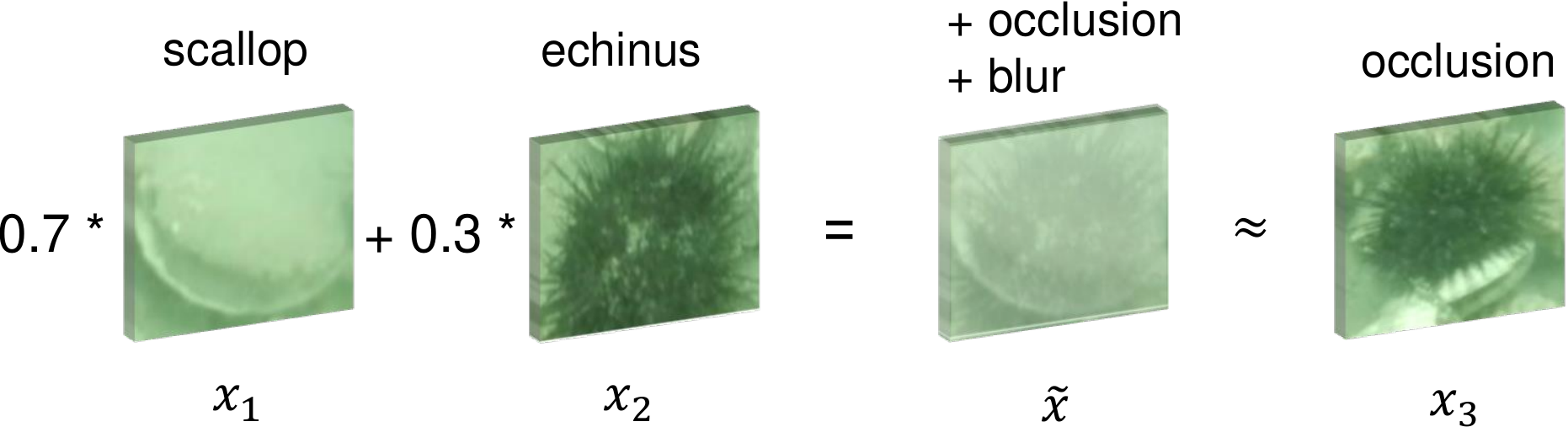}}
\end{minipage}
\caption{Visualization of RoIMix Method. $x_1$, $x_2$ are two RoIs containing a scallop and a sea urchin, respectively. $x_3$ is an occluded sample (a sea urchin lies on a scallop) cropped from an training image. Via RoIMix, $x_1$ and $x_2$ mix into a new virtual proposal $\widetilde{x}$ similar to $x_3$, simulating the situation of occlusion and blur.}
\label{fig:method}
\end{figure}

Using this method, we can get new virtual RoIs simulating overlapping, occluded and blurred objects. Figure \ref{fig:method} visualizes the process of our proposed method. We replace the original proposals with these new virtual RoIs and generate new training samples. We train the network by minimizing the original loss function on these generated samples. Code-level details are presented in Algorithm \ref{alg:roimix}.

\subsection{Discussion}
\label{ssec:subhead}
We simulate objects' overlap, occlusion by RoIMix to help the model implicitly learn better detection capability of dense objects. From the perspective of statistical learning theory, RoIMix is a type of linear interpolation between two proposals, and the decision boundary may become smoother without sharp transitions.

To be specific, RoIMix follows the VRM Principle instead of the ERM Principle, enabling deep learning models to be robust. A model trained following the ERM Principle minimizes empirical risk to help the model fit the training data well. We define empirical risk $R_\delta$ as
\begin{equation}
    R_\delta(f) =\frac{1}{n}\sum_{i=1}^nl(f(x_i), y_i),
\end{equation}
where $f$ represents the nonlinear expression between x and y, $n$ is the number of samples, and $l$ is a loss function measuring the distance between $f(x_i)$ and $y_i$. However, this training strategy makes the decision boundary fit the training data too much, and leads to over-fitting. Therefore, we suggest not using empirical risk to approximate the expected risk. RoIMix follows VRM rule and generates the vicinal distribution of training data. Then we can replace the training data $(x_i, y_i)$ with the vicinal data $(\widetilde{x}, \widetilde{y})$and approximate expected risk $R_v$:
\begin{equation}
    R_v(f)=\frac{1}{n}\sum_{i=1}^nl(f(\widetilde{x}), \widetilde{y}).
\end{equation}
Therefore, the training process is transformed to minimizing expected risk $R_v$. In each epoch, RoIMix generates different vicinal training data. In this manner, the model tends to be more robust. Session 4.3 illustrates the robustness of the model trained by RoIMix in detail. 

\begin{algorithm}[h]  
  \caption{RoIMix. The number of Images and RoIs in a mini-batch is N and n, respectively. RPN generates the same number of RoIs for each image. $\{RoI_i\}$ represents RoIs generated by RPN. $\{RoI_i\}$ corresponds to $\{RoI_j\}$ after random permutation of $\{RoI_i\}$. $(x_A, y_A)$, $(x_B, y_B)$ represents upper left corner and lower right corner of RoI. }  
  \label{alg:roimix}  
  \begin{algorithmic}[1] 
    \Require input images: $I \in \mathbf{R}^{N \times C \times H \times W}$, input RoIs: $\{RoI_i \in \mathbf{R}^{n \times c \times h \times w}\}$, $\{RoI_j \in \mathbf{R}^{n \times c \times h \times w}\}$, RoIs Position: ${\{x_{A_i}, y_{A_i}, x_{B_i}, y_{B_i}\}}_{i={0...n-1}}$
      \State initialize output image: $I' = I$
      \For{each $k$ in range(n)}
          \State choose two RoIs separately from $\{RoI_i\},\{RoI_j\}$: $x_i$, $x_j$
          \State generate mixing ratio $\lambda'$ using (2)(3)
          \State create mixed RoI $\widetilde{x}$ using (1)
          \State calculate the image index of $x_i$: $b = \frac{kN}{n}$
          \State paste generated RoI into image: $I'[b,\ :, y_{A_{i}}:y_{B_{i}}, x_{A_{i}}: x_{B_{i}}] = \widetilde{x}$
      \EndFor
    \Ensure new training sample $I'$
  \end{algorithmic}  
\end{algorithm}  

\begin{table*}[h!]
\centering
 \setlength{\tabcolsep}{0.4mm} {
 \begin{tabular}{c | c | c c c c c c c c c c c c c c c c c c c c} 
 \hline
 Method & mAP & areo & bike & bird & boat & bottle & bus & car & cat & chair & cow & table & dog & horse & mbike & person & plant & sheep & sofa & train & tv \\ 
 \hline\hline
 Baseline & 80.0 &\textbf{ 85.4} & \textbf{87.0} & \textbf{79.5} & 73.0 & 69.0 & 84.8 & 88.4 & 88.4 & 65.2 & 85.5 & 74.3 & 87.3 & 86.4 & 81.7 & 83.4 & 50.1 & 83.8 & \textbf{81.3} & 85.1 & 80.6\\
 \textbf{Proposed} & \textbf{80.8} & 85.3 & \textbf{87.0} & 79.1 & \textbf{73.9} & 70.2 & 86.9 & 88.3 & 88.8 &\textbf{ 66.0} & 86.1 & 75.1 & 88.2 & 88.0 & 85.6 & 83.1 &\textbf{ 54.8} & 83.8 & 81.1 & 86.3 & 79.0 \\
 \hline
  GTMix & 80.6 & 82.2 & 85.8 & 79.4 & 72.6 & \textbf{71.5} & 87.5 & \textbf{88.8} & 88.3 & 65.4 & \textbf{86.3} & \textbf{76.3} & 88.3 & \textbf{88.3} & 86.1 & \textbf{84.3} & 51.2 & 83.7 & 80.8 & 86.2 & 79.8\\
 Single\_GTMix & 80.5 & \textbf{85.4} & 86.2 & 78.7 & 72.4 & 69.7 & \textbf{88.2} & 88.4 & \textbf{89.0} & 65.4 & 85.2 & 73.3 & 87.3 & 87.8 & \textbf{86.2} & 83.0 & 53.0 & 81.2 & \textbf{81.3} & 85.6 &\textbf{ 82.8}  \\
 Single\_RoIMix & 80.3 & 80.8 & \textbf{87.0} & \textbf{79.5} & 72.3 & 69.2 & 87.3 & 88.5 & 87.9 & 64.1 & 86.0 & 74.2 & \textbf{88.7} & 87.3 & 84.9 & 83.0 & 54.7 & \textbf{84.5} & 79.0 & \textbf{86.4} & 80.1\\
 \hline
 \end{tabular}}
 \caption{Detection results on the VOC 2007 test set, trained on 07 trainval + 12 trainval.}
 \label{table:voc}
\end{table*}

\section{EXPERIMENT}
\label{sec:experiment}

\subsection{Experiments on URPC 2018}
\label{ssec:subhead}

We comprehensively evaluate our method on the URPC 2018. This dataset consists of 2901 trainval images and 800 test images over 4 object categories, including holothurian, echinus, scallop and starfish. We choose ResNet-101 pretrained on ImageNet as the backbone network and 128 RoI features are extracted from each input image. We use the default hyper-parameters for Faster R-CNN. Mean Average Precision (mAP) is adopted for the evaluation metric. In our experiments on URPC 2018, we set the hyper-parameter $a = 0.1$. 

The ablation study is shown in Table \ref{table:urpc}. Firstly, we directly generate mixing ratio by Eq.(2) without applying Eq.(3). The last two rows in Table \ref{table:urpc} show that the max operation brings 2.06\% and 1.8\% mAP gains, which illustrates the importance of Eq.(3). Secondly, we compare the effects of mixing Ground Truths (GTs) and mixing RoIs. The second to fifth rows in Table \ref{table:urpc} show that mixing RoIs contributes more to performance improvement than mixing GTs. Furthermore, we evaluate the importance of interactions among images. ``Single\_RoIMix'' means choosing and mixing proposals on a single image while our proposed method combines proposals from multiple images in the mini-batch. The second and the fifth rows in Table \ref{table:urpc} show that mixing RoIs among multiple images achieves 0.41\% mAP higher than mixing on a single image. Above all, the results show that the above variants lead to performance degradation compared to our proposed method.

\begin{figure}[htb]
\begin{minipage}[b]{.48\linewidth}
  \centering
  \centerline{\includegraphics[width=4.0cm, height=2cm]{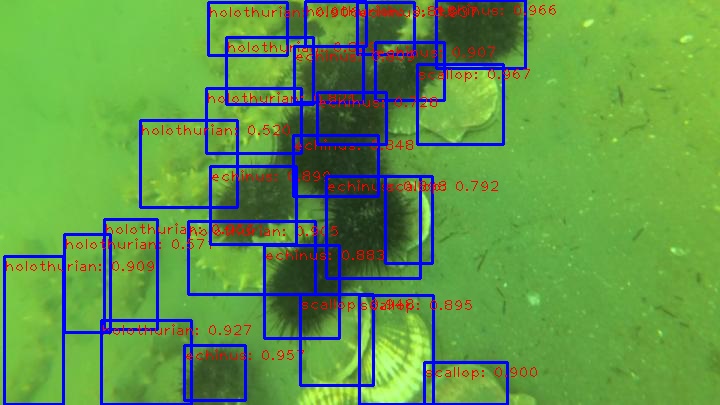}}
  \centerline{(a) baseline}\medskip
\end{minipage}
\hfill
\begin{minipage}[b]{0.48\linewidth}
  \centering
  \centerline{\includegraphics[width=4.0cm, height=2cm]{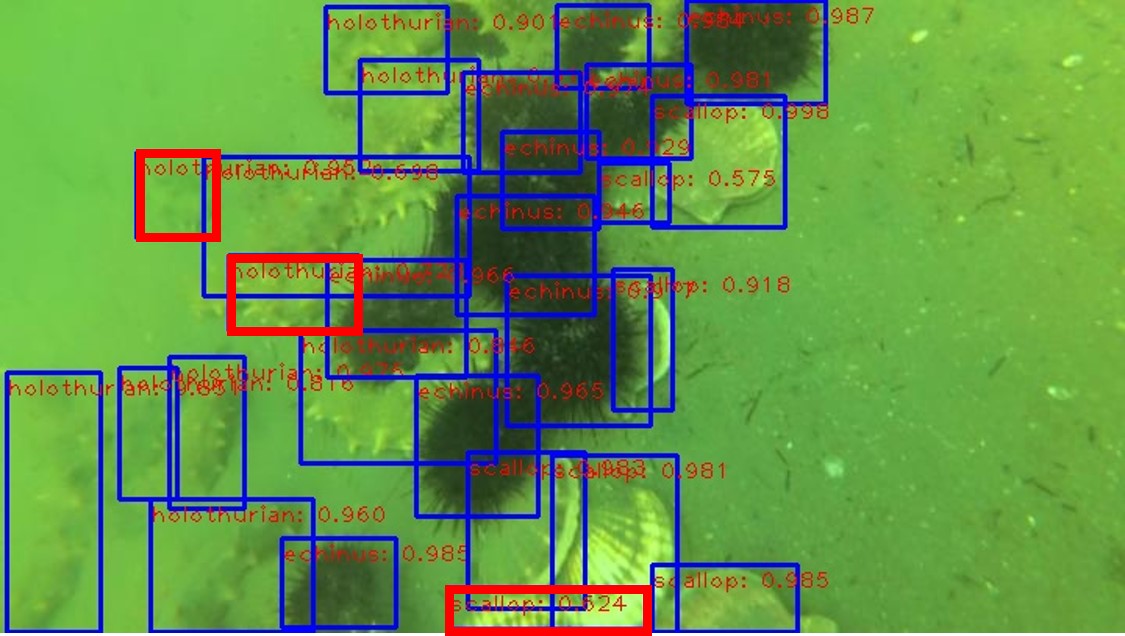}}
  \centerline{(b) RoIMix}\medskip
\end{minipage}
\caption{Comparison of detection results between baseline and our proposed method.}
\label{fig:det_res}
\end{figure}

Figure \ref{fig:det_res} visualizes the detection results of the baseline and our proposed method. There are three red boxes marked in Figure \ref{fig:det_res}(b), two of which are vague and overlapping holothurians, and the other is an incomplete scallop. The baseline model fails to detect the objects in the three red boxes, while our method is successful. This illustrates that our method has better detection capability of blurred, overlapping objects.

\subsection{Experiments on PASCAL VOC.}
\label{ssec:subhead}

We also evaluate the effectiveness of our proposed method RoIMix on generic object detection dataset Pascal VOC (07+12): the model is trained on a union set of VOC 2007 trainval and VOC 2012 trainval and tested on the VOC 2007 test set. We use the same setting as in the section 4.1. In the experiments on Pascal VOC, we set the hyper-parameter $a = 0.01$ empirically.

To our knowledge, it is the first experimental report for mixed-samples data augmentation on object detection. We compare our method with our baseline Faster R-CNN. Next we evaluate the performance of RoIMix variants. Table \ref{table:voc} shows that our proposed method achieves 0.8\% higher than the baseline and outperforms its variants. We observe that RoIMix's performance gain on Pascal VOC is smaller than on URPC. One possible reason is that there are more overlapping, occluded and blurred objects in the URPC, which is resolved by our method. Thus, the performance gain is larger on the URPC dataset.

\subsection{Stability and Robustness}
\label{ssec:subhead}

\begin{figure}[htb]
\begin{minipage}[b]{.48\linewidth}
  \centering
  \centerline{\includegraphics[width=4.0cm, height=3cm]{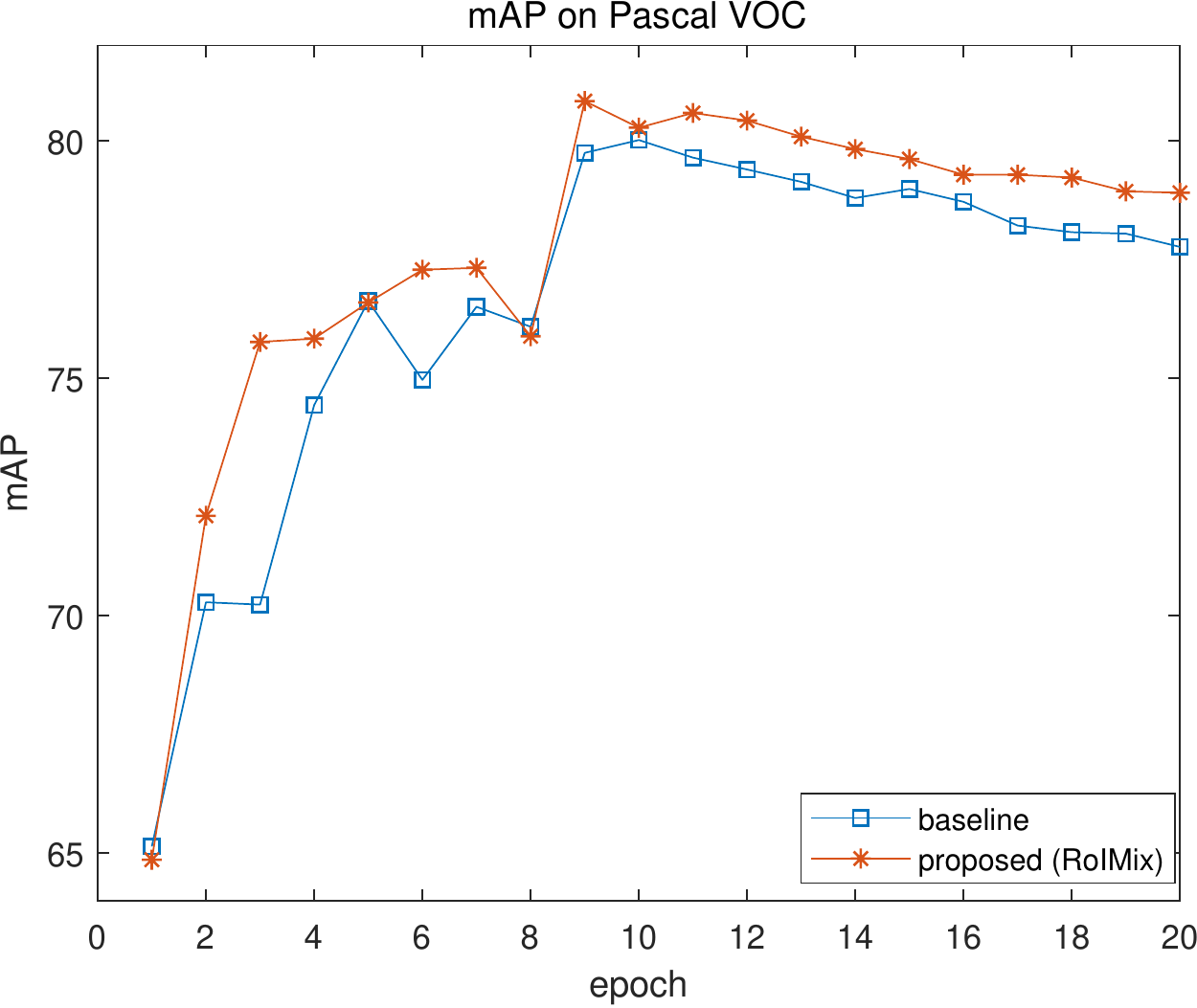}}
  \centerline{(a) Detection Result of VOC}\medskip
\end{minipage}
\hfill
\begin{minipage}[b]{0.48\linewidth}
  \centering
  \centerline{\includegraphics[width=4.0cm,height=3cm]{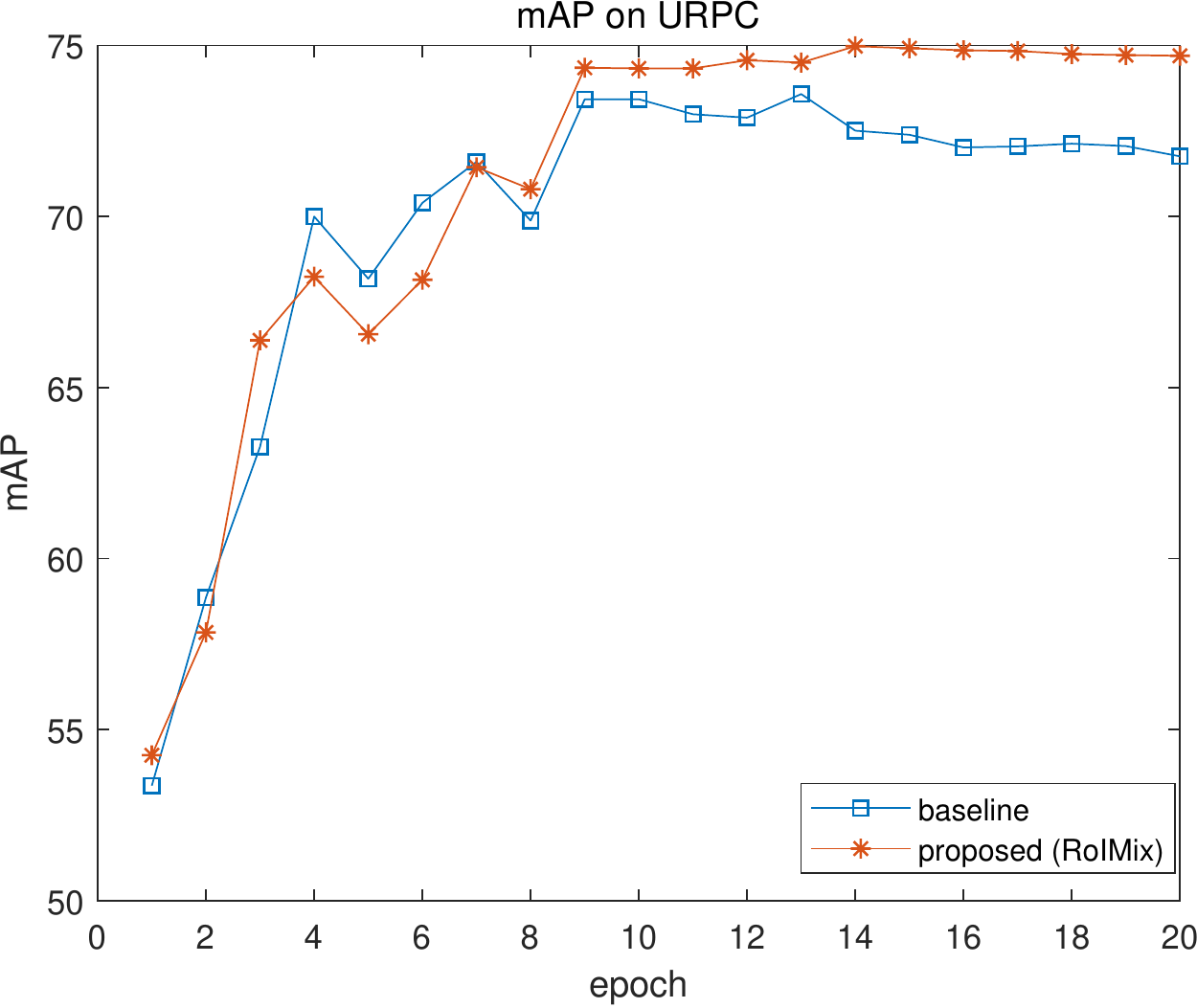}}
  \centerline{(b) Detection Result of URPC}\medskip
\end{minipage}
\caption{Analysis for stability of baseline and RoIMix. }
\label{fig:cmp_robust}
\end{figure}

We analyze the effect of RoIMix on stabilizing the training of object detectors. We report mean Average Precision (mAP) during the training with RoIMix against the baseline. We visualize the results on both Pascal VOC and URPC datasets in Figure \ref{fig:cmp_robust}.

First, we observe that RoIMix achieves much higher mAP than the baseline at the end of training in both datasets. After the mAP reaches its highest point, the baseline begins to face over-fitting with the increase of training epochs. On the other hand, RoIMix drops steadily in Pascal VOC and keeps its mAP curve better than the baseline over a large margin. In the URPC dataset, RoIMix remains stable as epochs increase after reaching the highest point of mAP. Furthermore, the maximum margin between our proposed method and baseline reaches 2.04\%. It shows that diverse vicinal training data generated by RoIMix can alleviate over-fitting and improve the stability of training process.

\begin{figure}[htb]
\begin{minipage}[b]{.48\linewidth}
  \centering
  \centerline{\includegraphics[width=4.0cm, height=2cm]{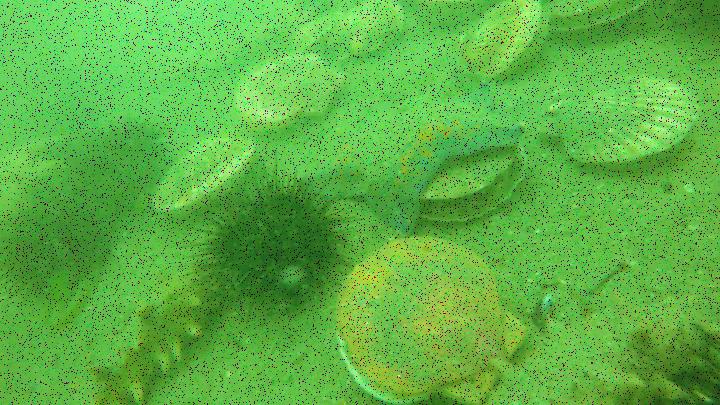}}
  \centerline{(a) Samples with Pepper Noise}\medskip
\end{minipage}
\hfill
\begin{minipage}[b]{0.48\linewidth}
  \centering
  \centerline{\includegraphics[width=4.0cm, height=2cm]{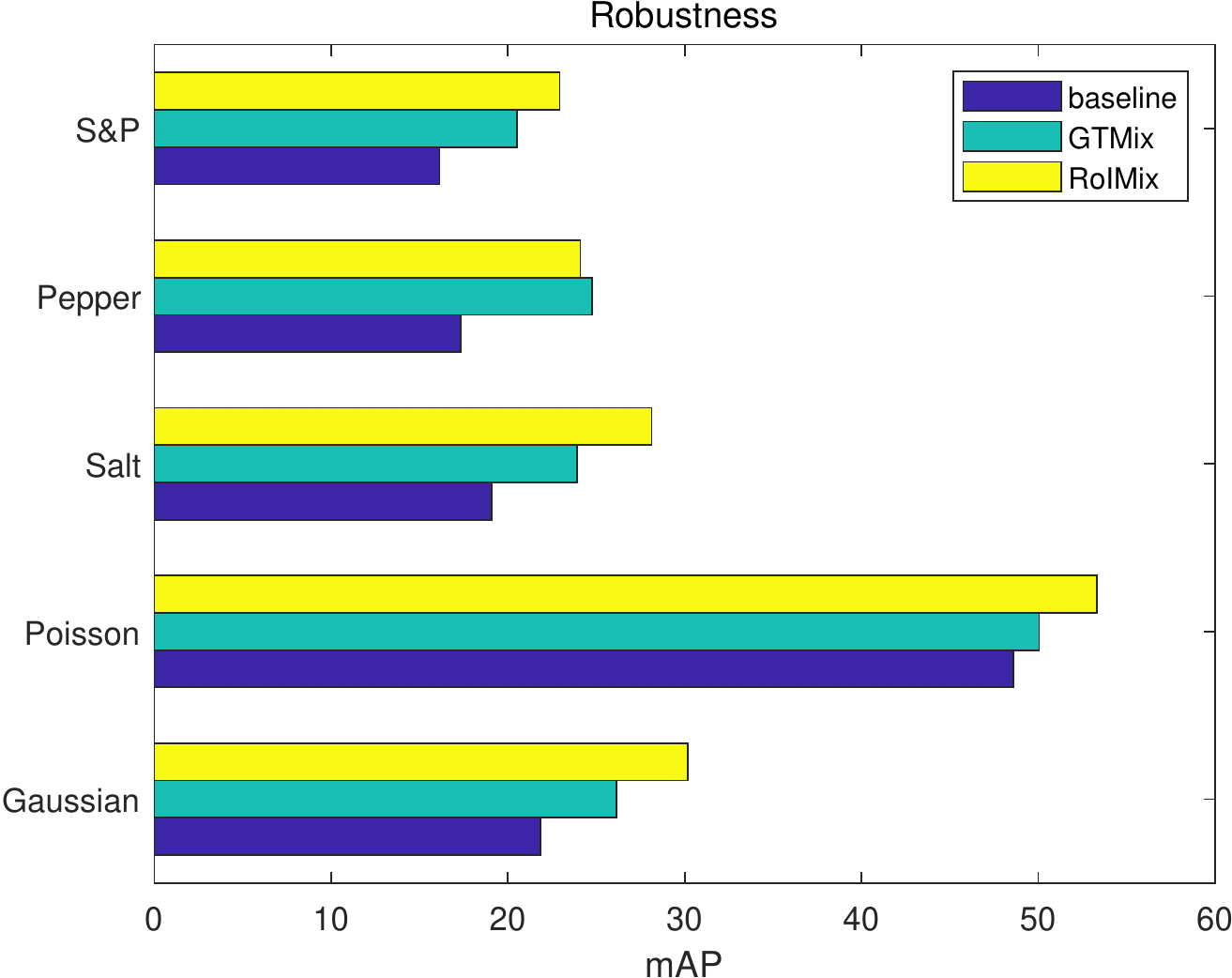}}
  \centerline{(b)Analysis for different noise}\medskip
\end{minipage}
\caption{Robustness experiments on the URPC 2018. }
\label{fig:robust}
\end{figure}

Furthermore, we evaluate the robustness of the trained model by applying 5 types of artificial noise samples: Gaussian noise, Poisson noise, salt noise, pepper noise, and salt-and-pepper noise. Figure \ref{fig:robust}(a) displays the sample with pepper noise. We use ImageNet pre-trained ResNet-101 model with same setting as in Section 4.1. We evaluate the baseline, GTMix, and RoIMix on each type of noise samples and visualize the results in Figure \ref{fig:robust}(b). The maximum performance gap between our proposed method and the baseline among these 5 types of noises is 9.05\% mAP. The histogram shows that our proposed method is more robust against noise perturbations.

Apart from artificial noise samples, we additionally explore the situation of vagueness by applying Gaussian Blur to the test images. As shown in Table \ref{table:blur}, we can see that the performance is improved by 0.7\% mAP after adopting the RoIMix method. These experiments further illustrate that RoIMix results in better robustness.

\begin{table}[h!]
\centering
 \setlength{\tabcolsep}{1mm} {
 \begin{tabular}{c | c | c | c c c c} 
 \hline
  Method & mAP & Delta & holothurian & echinus & scallop & starfish \\
  \hline
  Baseline &70.47 & 0 & 64.36 & \textbf{86.54} & \textbf{48.83} & 82.16\\ 
  GTMix &70.70 & +0.23 & 65.92 & 86.13 & 48.09 & 82.66\\
  \textbf{Proposed} & \textbf{71.17} & \textbf{+0.70} &\textbf{66.44} & 86.17 & 48.72 & \textbf{83.35}\\

 \hline
 \end{tabular}}
 \caption{Detection results on artificial Gaussian Blur samples. We apply baseline, GTMix and RoIMix methods on these blur samples. Delta represents their performance gains with respect to the baseline.}
 \label{table:blur}
\end{table}

\section{CONCLUSION}
\label{sec:conclude}

In this paper, we propose RoIMix for underwater object detection. To the best of our knowledge, it is the first work to conduct proposal-level fusion among multiple images for generating diverse training samples. RoIMix aims to simulate overlapping, occluded and blurred objects, enabling the model implicitly learn the capability of detecting underwater creatures. The experiments show that our proposed method can improve the performance on URPC by 1.18\% mAP and on Pascal VOC by 0.8\% mAP. Besides, RoIMix exhibits more stability and robustness. RoIMix was used in our first-prize solution for URPC2019 offline target recognition.

\vfill\pagebreak

\bibliographystyle{IEEEbib}
\bibliography{refs}

\end{document}